\documentclass[letterpaper]{article}
\usepackage{uai2019}
\usepackage[margin=1in]{geometry}

\usepackage{amsmath}
\usepackage{graphicx}
\usepackage{multirow}
\usepackage{float}
\usepackage{amsfonts}
\usepackage{amsthm}
\usepackage[ruled,vlined]{algorithm2e}
\usepackage{tikz}
\usepackage{amssymb}
\usepackage{makecell}
\usetikzlibrary{fit, arrows, arrows.meta}
\usepackage{booktabs}
\usepackage{hyperref}
\usepackage{paralist}
\usepackage{subcaption}

\newtheorem{defi}{Definition}

\usepackage{times}

\title{Synthesizing Unrestricted False Positive Adversarial Objects \\ Using Generative Models}

\author{ {\bf Martin Kotuliak} \\
ETH Zurich \\
komartin@student.ethz.ch\\
\And
{\bf Sandro E. Sch\"{o}nborn}  \\
ABB Future Labs \\
sandro.schoenborn@ch.abb.com \\
\And
{\bf Andrei Dan}   \\
ABB Future Labs \\
andrei.dan@ch.abb.com   \\
}

\begin{document}

\maketitle

\begin{abstract}
% Adversarial examples
Adversarial examples are data points misclassified by neural networks. Originally, adversarial examples were limited to adding small perturbations to a given image. Recent work introduced the generalized concept of unrestricted adversarial examples, without limits on the added perturbations.
% This work
In this paper, we introduce a new category of attacks that create unrestricted adversarial examples for object detection. Our key idea is to generate \emph{adversarial objects} that are \emph{unrelated} to the classes identified by the target object detector. Different from previous attacks, we use off-the-shelf Generative Adversarial Networks (GAN), without requiring any further training or modification. Our method consists of searching over the latent normal space of the GAN for adversarial objects that are wrongly identified by the target object detector. We evaluate this method on the commonly used Faster R-CNN ResNet-101, Inception v2 and SSD Mobilenet v1 object detectors using logo generative iWGAN-LC and SNGAN trained on CIFAR-10. The empirical results show that the generated adversarial objects are indistinguishable from non-adversarial objects generated by the GANs,  transferable between the object detectors and robust in the physical world. This is the first work to study unrestricted false positive adversarial examples for object detection.

\end{abstract}

\section{INTRODUCTION}

% flatex input: [ini_fig.tex]
\begin{figure*}
\centering
    \begin{tikzpicture}[every text node part/.style={align=center}]
    
    \newsavebox{\picbox}

    \tikzstyle{vector} = [rectangle, minimum width=1cm, minimum height=0.2cm,text centered, draw=black, fill=red!30, rotate=90,anchor=north]
    \tikzstyle{NN} = [rectangle, rounded corners, minimum width=3cm, minimum height=1cm,text centered, draw=black, fill=blue!30]
    \tikzstyle{detector} = [rectangle, rounded corners, minimum width=10cm, minimum height=7cm,text centered, line width=2mm, draw=blue!20, label={[shift={(0cm,-1cm)}, color=red]north:#1}]
    \tikzstyle{arrow} = [color=black!30, line width = 3pt,->,>=angle 60]
    \tikzstyle{arrow2} = [color=red!30, line width = 3pt,->,>=angle 60]
    
    \draw[color=red, fill=red] (5pt,0) circle (5pt);
    \draw[color=red, fill=red] (5pt,13pt) circle (5pt);
    \draw[color=red, fill=red] (5pt,26pt) circle (5pt);
    \draw[color=red, fill=red] (5pt,-13pt) circle (5pt);
    \draw[color=red, fill=red] (5pt,-26pt) circle (5pt);
    \node (z) at (5pt, 45pt) {Latent \\ variables};

    \draw [fill=black!30, line width=1.5pt] (2.9, -1) -- (2.9, 1) -- (3.9, 1.3) -- (3.9,-0.7) -- (2.9, -1);
    \draw [fill=black!30, line width=1.5pt] (2.5, -1) -- (2.5, 1) -- (3.5, 1.3) -- (3.5,-0.7) -- (2.5, -1);
    \draw [fill=black!30, line width=1.5pt] (2.1, -1) -- (2.1, 1) -- (3.1, 1.3) -- (3.1,-0.7) -- (2.1, -1);
    \node (gan) [rectangle, fill=black!30] at (3.1,0) { GAN};
    \node at (3.1,1.7) {Logo generator};
    
    \savebox{\picbox}{\includegraphics[width=1.5cm]{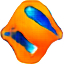}}
    \node (logo) [rounded corners=2pt, minimum width=\wd\picbox,
        minimum height=\ht\picbox, path picture={
         \node at (path picture bounding box.center) {
            \usebox{\picbox}};
        }] at (6.3, 0) {};
        
    \node (logo_label) [yshift=27pt] at (logo.north) {Adversarial object};
        
    \draw [arrow] (0.6, -0.3) -- (2, -0.3);
    \draw [arrow2] (2, 0.3) -- (0.6, 0.3);   
    
    \draw [arrow] (4, -0.3) -- (5.2, -0.3);
    \draw [arrow2] (5.2, 0.3) -- (4, 0.3);

    \savebox{\picbox}{\includegraphics[width=3cm]{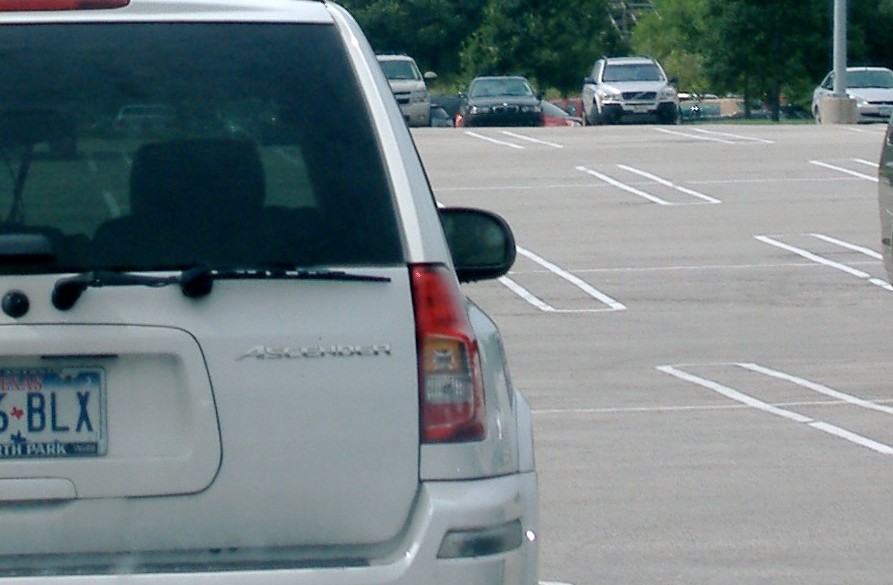}}
    \node (bckg) [rounded corners=2pt, minimum width=\wd\picbox,
        minimum height=\ht\picbox, path picture={
         \node at (path picture bounding box.center) {
            \usebox{\picbox}};
        }] at (1.4, -3.5) {};
        
    \node (bckg_label) [yshift=-17pt] at (bckg.south) {Input image};

    \draw [arrow] (12, -0.3) -- (13.5, -0.3);
    \draw [arrow2] (13.5, 0.3) -- (12, 0.3); 
    \node at (15, 0.3) {Gradient flow};
    \node at (15, -0.3) {Data flow};

    \savebox{\picbox}{\includegraphics[width=3cm]{back_of_car.jpg}}
    \node (bckg2) [rounded corners=2pt, minimum width=\wd\picbox,
        minimum height=\ht\picbox, path picture={
         \node at (path picture bounding box.center) {
            \usebox{\picbox}};
        }] at (6.3, -3.5) {};
    
    \node (bckg_logo) [yshift=-17pt] at (bckg2.south) {Input image overlaid \\  with adversarial object};
    
    \draw [arrow] (3,-3.5) -- (4.7,-3.5);
    % \draw [arrow2] (5.4,-3.2) -- (2.6,-3.2);
    
    \draw [arrow] (6, -1.1) -- (6, -2.3);
    \draw [arrow2] (6.6, -2.3) -- (6.6, -1.1);    
        
    \savebox{\picbox}{\includegraphics[width=17pt]{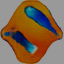}}
    \node (logo2) [rounded corners=0pt, minimum width=\wd\picbox,
        minimum height=\ht\picbox, path picture={
         \node at (path picture bounding box.center) {
            \usebox{\picbox}};
        }, xshift=-12pt, yshift=-11pt] at (bckg2.center) {};

    % \savebox{\picbox}{\includegraphics[width=3cm]{NeuralNetwork.png}}
    % \node (det) [rounded corners=2pt, minimum width=\wd\picbox,
    %     minimum height=\ht\picbox, path picture={
    %      \node at (path picture bounding box.center) {
    %         \usebox{\picbox}};
    %     }, label={[]north:Detector}] at (11, -3.5) {};
        
    \draw [fill=black!30, line width=1.5pt] (10.4, -4.5) -- (10.4, -2.5) -- (11.4, -2.2) -- (11.4,-4.2) -- (10.4, -4.5);
    \draw [fill=black!30, line width=1.5pt] (10, -4.5) -- (10, -2.5) -- (11, -2.2) -- (11,-4.2) -- (10, -4.5);
    \draw [fill=black!30, line width=1.5pt] (9.6, -4.5) -- (9.6, -2.5) -- (10.6, -2.2) -- (10.6,-4.2) -- (9.6, -4.5);
    \node (det) [rectangle, fill=black!30] at (10.56,-3.5) {DNN};
    
    \node at (10.52,-5.1) {Traffic Sign \\ Detector};
    
    \draw [arrow] (7.9,-3.8) -- (9.5,-3.8);
    \draw [arrow2] (9.5,-3.2) -- (7.9,-3.2);   
        
    \savebox{\picbox}{\includegraphics[width=3cm]{back_of_car.jpg}}
    \node (bckg3) [rounded corners=2pt, minimum width=\wd\picbox,
        minimum height=\ht\picbox, path picture={
         \node at (path picture bounding box.center) {
            \usebox{\picbox}};
        }] at (14.3, -3.5) {};
        
    \draw [arrow] (11.45,-3.8) -- (12.8, -3.8);
    \draw [arrow2] (12.8, -3.2) -- (11.45,-3.2) ;
        
    \savebox{\picbox}{\includegraphics[width=17pt]{logo.jpg}}
    \node (logo3) [rounded corners=0pt, minimum width=\wd\picbox,
        minimum height=\ht\picbox, path picture={
         \node at (path picture bounding box.center) {
            \usebox{\picbox}};
        }, xshift=-12pt, yshift=-11pt] at (bckg3.center) {};
        
    \draw[color=green, line width = 2pt] (logo3.south west) rectangle (logo3.north east);
    \node[fill=green, yshift=6pt, xshift=-8pt] at (logo3.north east) {\tiny 99\%};
    
    \node (bckg_label3) [yshift=-17pt] at (bckg3.south) {Adversarial object  \\ detected as traffic sign};
        
    \end{tikzpicture}
    
    \caption{Overview of our method for creating adversarial objects using off-the-shelf GAN.}
    \label{fig:proc}
\end{figure*}
% flatex input end: [ini_fig.tex]

% This work

Deep Neural Networks (DNN) are increasingly used in safety critical applications. One important use case is using machine learning for traffic sign detection. With the popularity of DNN models used in critical applications, the threat of creating malicious imagery also increases. These attacks are called adversarial examples. The role of an adversarial example is to fool the DNN in classifying the input image differently from how a human would.

% adversarial examples, hidden to the human observer
Neural networks have been shown to be vulnerable to adversarial examples in \cite{szegedy2013intriguing}.  Follow up work \cite{carlini2017towards, FGSM, moosavi2016deepfool, su2019one} improved the algorithms to generate adversarial examples based on small perturbations of the input. \cite{song2018unconstrained, wang2019nonconstrained} use generative adversarial networks (GAN) to create unrestricted adversarial examples, instead of perturbing existing images.
% We build on their ideas to create a powerful attack in physical-world.
%
% introduce our new approach
While many techniques focus on digital attacks, where the attacker has access to either the camera or the IT infrastructure, it is important to explore and understand feasible physical attacks in more detail.

In this work, we introduce a new attack that synthesizes unrestricted adversarial examples for object detection. We believe this is the first work that investigates unrestricted adversarial examples that cause false positives for object detection. Our method generates, using off-the-shelf GANs, adversarial objects that are wrongly identified by object detectors. We evaluate our attack in a physical world setting and observe that the adversarial objects are robust to the positioning of the camera.

%which will be attributed to a mistake in DNN model and not to a malicious activity. People often put logos of companies, bands, or political groups on their cars, buildings, shops. Therefore creating a malicious logo attacking a traffic sign DNN detector would not raise suspicion. With advent of deep fake videos, photos, or voice recording generated by GANs we can see how easy it is to create artificial imagery.

% describe the approach (figure)

\autoref{fig:proc} shows an overview of the new attack, instantiated for traffic sign detectors and using a GAN that generates logos. Since logos are frequently found on cars, buildings, shops, billboards, a malicious logo that attacks a traffic sign DNN detector has small changes of raising suspicion. Our algorithm optimizes latent variables, such that the GAN synthesizes logos that look natural and are adversarial to the target traffic sign detector. In this example, the logo is detected as a danger sign by the traffic sign DNN detector.

% contributions
\paragraph{Our Contributions} The main contributions of this work are:

\begin{enumerate}
    \item We introduce a new attack category, namely unrestricted false positive adversarial objects, that are unrestricted adversarial examples for object detection which resemble other natural objects in physical world.

    \item The paper describes a novel and effective algorithm for generating adversarial objects for object detection neural networks, using pre-trained GANs.

    \item We implement our algorithm and evaluate it using widely used Faster R-CNN Object Detection networks based on ResNet-101 and Inception v2 and the SSD Mobilenet, trained for traffic sign detection. To generate the adversarial objects, we evaluate two generative models: logo generating iWGAN-LC and the SNGAN trained on CIFAR-10.

    \item We demonstrate the transferability of the synthesized adversarial examples between different object detection networks and the robustness of the attacks in a physical setup.

\end{enumerate}

In the next section, we provide background information on adversarial examples. Next, in \autoref{sec:des} we explain the theory behind our attack and we present the results of our empirical evaluation in \autoref{sec:eval}. Before concluding, we present related work in \autoref{sec:related}.

\section{BACKGROUND} \label{sec:background}

\subsection{Adversarial Examples}

Let $\mathcal{I}$ be the set of all digital images and $\mathcal{O} \subseteq \mathcal{I}$ a subset of natural images from a specific domain (e.g., handwritten numbers). Oracle $o: \mathcal{I} \longrightarrow \{1,2,...K, \nu, \phi\}$ maps images in its domain $\mathcal{O}$ to $K$ classes, all other natural images to a class $\nu$, and unnatural (noisy) images to class $\phi$. The oracle $o$ represents a human observer. The classifier $f:\mathcal{I} \longrightarrow \{1,2,...K\}$ maps all digital images to one of $K$ classes. As defined in \cite{song2018unconstrained}, a \emph{perturbation-based adversarial example} $x$ is an image $x$ obtained by perturbing an original image $x'$ such that the classifier $f$ maps the original and the perturbed images to different classes. The perturbation must be small enough to fool the classifier but not the oracle.

%adding a small perturbations to an original image, such that the classifier maps the original image and the perturbed image to different classes.

\begin{defi}[Perturbation-Based Adversarial Examples]

Perturbation-based adversarial examples $\mathcal{A}_p$ are defined for a (test) subset of images $\mathcal{T} \subseteq \mathcal{O}$, and a small perturbation bound $\epsilon$ as:

$\mathcal{A}_p \triangleq \{x \in \mathcal{O} \; | \; \exists x' \in \mathcal{T}: \; ||x - x'|| \leq \epsilon \wedge f(x) \neq f(x') = o (x) = o (x') \}$
\end{defi}

To generate a perturbation-based adversarial example for an object classifier, one must solve an optimisation problem. \cite{yuan2019adversarial} defines a general optimisation problem:
\begin{align*}
    \text{min}_{x'} & ||x'-x||\\
    \text{s.t.} & f(x') \neq f(x)\\
    & x \in [0, 1]
\end{align*}
where $||\cdot||$  measures the distance between the adversarial example and the original image. Large perturbations might fool both the human and the classifier, making the attack fail the task, or make the adversarial example be outside of intended domain $\mathcal{O}$.

The distance function between the original image and the adversarial example measures how close the images are to each other. An ideal distance function would model human perception, rather than the pixel-based distance. \cite{rozsa2016adversarial} tried to create a more complex function to capture human vision through luminance, contrast, and structural measures. \cite{jang2017objective} based the similarity function on Fourier transformations around the edges. In practice, most research work uses the $l_p$ distance as a proxy.

%and takes as input the original image and the adversarial image. The function would measure how close they are to human perception. Because two images with close pixel values might look to a human very distant, and vice versa. Most of the research work uses the $l_p$ distance as a proxy.

%This general problem is hard to optimize due to high-non-linearity of $||\cdot||$ function \cite{carlini2017towards}. There are many ways how to transform the problem, i.e. put distance of adversarial example into the constraints and the difference of outputs into the objective. \cite{yuan2019adversarial} presents a  summary of the various approaches.

The optimization of the general problem is hard due to the high non-linearity of the distance functions \cite{carlini2017towards}. For a summary on different approaches to reformulating the problem, we refer the reader to \cite{yuan2019adversarial}.

\cite{carlini2017towards} solved the optimisation problem for perturbations $\eta$ on top of original image $x'$:
\begin{align}
\label{eq:carliniwagneropt}
    \text{min}_\eta \, & \left || \eta \right ||_p + \kappa g(x'+\eta, x') \\
    \text{s.t.} \, & x' + \eta \in [0,1]^n
\end{align}
in an iterative manner, using gradient descent. The function $g$ evaluates classification error of the target classifier. Value of $g$ for an original and an adversarial image is low they are classified to different classes, and high if they are classified as the same class. The $l_p$ norm $||\eta||_p$ regularizes the amount of perturbation applied. $\kappa$ is a unitless trade-off parameter for regularization. Finally, the constraint enforces the image with perturbation is still a valid picture with pixel values in the range of $[0, 1]$.

 \subsection{Generative Adversarial Network}

A Generative Adversarial Network (GAN) is a network consisting of a discriminator and a generator. The discriminator's goal is to distinguish generated images from real ones. The generator takes as an input a latent variable $z \sim \mathcal{N}(0, 1)$ and outputs an image. Its goal is to fool the discriminator to not recognize that the image is artificial. This goal is expressed as a loss function during the training phase of the GAN. GANs can be conditioned to generate images in a specific class $c$:  $\operatorname{GAN}(z, c)$.

\subsection{Unrestricted Adversarial Examples}

\cite{song2018unconstrained} explores the idea of creating adversarial images that are not constrained by a distance function. The restrictions for adversarial images are lifted, and they only require the classification to be different from the oracle label.

\begin{defi}[Unrestricted Adversarial Examples]
\cite{song2018unconstrained} defines unrestricted adversarial examples as $\mathcal{A}_u \triangleq \{x \in \mathcal{O} \; | \; \; f(x) \neq o (x) \}$
\end{defi}

They train a GAN network that approximates the domain $\mathcal{O}$ of images. The GAN is further conditioned on the class $c$ of images to generate. Assuming that a perfect GAN network generates natural images of class $c$ if $z$ is sampled from the GAN's latent distribution $Q$: $\forall z \sim Q, c:\operatorname{GAN}(z, c) \in \mathcal{O} \wedge o(\operatorname{GAN}(z, c)) = c$, they find latent variables $z$ such that $f(\operatorname{GAN}(z, c)) \neq c$. The optimization problem is then simplified as:
\begin{align*}
    \text{min}_z \; g'(\operatorname{GAN}(z, c), c)
\end{align*}
for a new classification error function $g'$. $g'$ evaluates to a low value for a latent variable $z$ and a target class $c$, if an image $\operatorname{GAN}(z, c)$ is not classified as $c$. It is evaluated to a greater value if it is classified as $c$. They overcome issues of using distance functions to measure perturbations by using GAN generated images.
% \begin{align*}
%     \forall z_1, z_2, c \; | \; f(\operatorname{GAN}(z_1, c)) \neq c \wedge f(\operatorname{GAN}(z_1, c)) = c: \\ g'(\operatorname{GAN}(z_1, c), c) < g'(\operatorname{GAN}(z_2, c),c)
% \end{align*}

\subsection{White-box vs Black-box attack}

\cite{yuan2019adversarial} classifies the attacks in two categories depending on the attacker model. A white-box attacker has full access to the attacked DNN, can view individual trained parameters and perform operations with it.
On the other hand, a black-box attack limits the access to the target DNN. The attacker is only allowed to feed images into the target network and observe outputs and cannot directly observe the inner working of the model.

\subsection{Adversarial Objects}

When a physical world scenario is considered, the attacker can not perturb the entire scene. Most of the physical world attacks are conducted using adversarial stickers. An adversarial sticker is a printable 2D shape with a filling that is chosen by the attacker \cite{sharif2016accessorize, eykholt2018classification, eykholt2018detection, chen2018shapeshifter}.

In computer vision, we discriminate the models between classifiers and object detectors. A classifier outputs the class of an image, whereas detectors label objects in the digital image. \cite{lu2017standard} showed that attacking an object detector is more difficult than attacking an object classifier. The object detector finds an object in the image whereas the object classifier only outputs the class of the image. An adversarial object is an adversarial example attacking the object detector.

% Out of the four physical attack papers mentioned above, only \cite{eykholt2018detection, chen2018shapeshifter} focus on attacks against object detectors, whereas the other two focus on object classifiers.

Successful adversarial examples are either classified as false positives (FP) or false negatives (FN). A false negative attack occurs when a perturbed input is classified differently from the input without perturbation, e.g. ignored or incorrectly classified. A false positive is when a input which would not be classified at all is assigned a class. A FP attack makes object detectors detect at least one object of the target class, even though it is not present.

% The aforementioned papers all explore false negative attacks.

\subsection{Object Detectors}
Faster R-CNN \cite{ren2015faster} is a neural network architecture for object detection. It first generates generic object bounding box proposals, which it then classifies in the second step. The network is optimized for speed, by using only a single CNN for both stages.

The Single Shot Detector \cite{liu2016ssd} uses one propagation through the network to detect all objects and assign them classes. It detects a high number of bounding boxes in different sizes, and then picks one bounding box for each object.

\section{Design}\label{sec:des}

We present our novel attack that generates unrestricted false positive  adversarial objects for object detection.

\subsection{Unrestricted False Positive Adversarial Examples}

The oracle $o$ not only labels objects in domain $\mathcal{O}$ to $K$ classes, as defined in \autoref{sec:background}, it also differentiates between all other natural images $\nu$ and noisy, unnatural digital images $\phi$. A detector $d$ works similarly to the classifier $f$, however it has an extra class $\bot$ for all undetected images. These correspond to images that the detector does not identify as being one of the $K$ target classes, but as other natural or noisy images (classes $\nu$ and $\phi$ of the observer).

\begin{defi}[Unrestricted False Positive Adversarial Examples] Unrestricted FP adversarial examples are defined as $\mathcal{A}_{fp} \triangleq \{x \in \mathcal{I \setminus O} \; | \; o(x)=\nu \wedge d(x)\in \{1,2,...K\} \}$.
\end{defi}

$\mathcal{A}_{fp}$ is a set of natural images (such that they can occur in physical world) that a detector identifies as one of the $K$ classes, but to a human observer they look like one of many objects in physical world, not part of one of the $K$ categories.
Finally, we define the set $\mathcal{M} = \{x \in \mathcal{I \setminus O} \; | \; o(x)=\nu\}$ containing all the natural images outside the $K$ target classes.

\subsection{Attacker model}

The attacker model that we consider assumes that he can only augment the physical world. However, to generate the proposed adversarial object, the attacker requires full access to the object detection network (white-box). Our empirical evaluation in \autoref{sec:eval} shows the transferability of adversarial objects between different object detection models trained on the same dataset. This allows the attacker to perform an attack without having access to the exact target object detector (black-box).

\subsection{Effective Generation of Unrestricted False Positive Adversarial Objects}

To generate an object in $\mathcal{M}$, we choose an arbitrary unconditional GAN such that $\forall z: \operatorname{GAN}(z) \in \mathcal{M} \, \wedge \, \exists z: d(\operatorname{GAN}(z))=\{1,2,...K\}$. Such a GAN defines a subset of $\mathcal{M}$. The attacker can select which domain to use by using different GANs.

We optimize for the latent vector $z$ in a loss function $\mathcal{L}$ which is composed of two terms:
\begin{align}\label{eq:opt}
\mathcal{L}=\mathcal{L}_0 + \kappa \mathcal{L}_1
\end{align}
$\mathcal{L}_0$ is a term enforcing an adversarial output, $\mathcal{L}_1$ is enforcing the natural look of the GAN output and $\kappa$ is a unitless trade-off parameter for regularization.

%\begin{align*}
%    \mathcal{L'}_0 = - \log \; \text{Pr}[d(\operatorname{GAN}(z)) \in \{1,2,...K\}]
%\end{align*}

%The $\mathcal{L'}_0$ term is forcing the optimizer find a latent vector $z$ such that the GAN generates an image which is classified as an object from the domain $\mathcal{O}$.

To perform a targeted attack for a specific class $c$ within $\mathcal{O}$ we can use the loss term:

%We can simply ange $\mathcal{L'}_0$ to perform a targeted attack for a specific class $c$ within $\mathcal{O}$:
\begin{align*}
    \mathcal{L'}_0 = - \log \; \text{Pr}[d(\operatorname{GAN}(z)) = c]
\end{align*}

$\mathcal{L'}_0$ is forcing the optimizer to find a latent vector $z$, such that the GAN generates an image which is classified with the target class label $c$.

An object detection DNN performs both object proposal generation and detection. It takes as input a digital image $x'$ and creates $n$ object proposals that the object detection DNN should identify. Proposals $B_1, B_2, ... B_n$ correspond to cropped images of objects in the original image. $B_k(x')$ is the $k^{\text{th}}$ object in $x'$. Since we do not know which proposal will correspond to our adversarial object, $\mathcal{L'}_0$ must be reformulated as:

\begin{align}\label{eq:l0}
    \mathcal{L}_0 = - \sum_{i=1}^n \log \text{Pr}[d(B_i(\operatorname{GAN}(z) \oplus x')) = c]
\end{align}

To generate a physical adversarial object, we first need to simulate it in a digital image and then print it. The $\oplus$ operation signifies the overlap of the generated object over a small space of original image.

A positive outcome of the new loss function is that the object detection DNN can detect multiple objects in the same adversarial object, increasing the strength of the attack.

%Given there does not exist a perfect GAN such that $\forall z: \operatorname{GAN}(z) \in \mathcal{M}$, we must enforce $z$ to have its intended distribution.

For a GAN to create samples approximately from the training distribution, we need to ensure that $z$ follows its intended distribution $z \sim Q$. This is the distribution which was used during training of the generator. For $z$ outside of $Q$, the generated images look unnatural. For many available GAN networks, $Q$ is a standard normal distribution $z \sim \mathcal N(0, I)$. Although we restrict ourselves to normal distributions in this work, it is important to note that the same procedure analogously applies to more complex distributions of $z$.

We implement the regularization loss $\mathcal{L}_1$ based on the difference between the intended and the estimated distribution of $z$. In order to quantify the difference between two distributions, we use a KL divergence:
\begin{align}\label{eq:kl}
    \mathcal{L}_1 = D_{KL}(P||Q)
\end{align}
where $P$ is the estimated distribution of current $z$ and $Q$ is its intended distribution.

Given the intended distribution of $z \sim \mathcal{N}(0,I)$, we initialize $z$ by sampling each element from $\mathcal{N}(0,1)$. To estimate $z$ at each iteration, we assume that each element is drawn from the same normal distribution. We estimate its mean $\hat{\mu}_P$ and variance $\hat{\sigma}_P^2$ by computing them over all elements of $z$. For $Q = \mathcal{N}(0,1)$:

\begin{align}
     D_{KL}(P||Q) = \log \frac{1}{\hat{\sigma}_P} + \frac{\hat{\sigma}_P^2 + \hat{\mu}_P^2}{2} - \frac{1}{2}
\end{align}

% Contrary to Carlini \& Wagner's attack, we do not need to constrain the final image to be in $[0, 1]$ range. If the object-generating GAN creates valid images we do not risk out-of-range values by addition since the pixel values are replaced by the object's values.

% In order to attack detectors and not only classifiers we had to change the function $g$. We perform a False Positive attack by covering parts of the original image with a new adversarial sticker. We optimize the sticker GAN$(z)$ to be recognized as class $c$. For this purpose \cite{chen2018shapeshifter} proposes the function $g$ to cover probability of detection by classifier C in all available detection boxes ($B_1, B_2, \ldots, B_N$, e.g. for R-CNN)

%Since the adversarial sticker does not occur in the original image $x$, we are performing a False Positive attack. We optimise for $Gan(z)$ sticker to be recognized as class $c$. One of R-CNN parameters is number of detection boxes ($B_1, B_2, ... B_n$) it can find. \cite{chen2018shapeshifter} proposes following:

% \begin{align*}
%     g(x, c) = - \sum_n \text{Pr}[\operatorname{C}\left (B_i(x) \right ) = c].
% \end{align*}

% \begin{figure}[h]
%     \centering
%     \includegraphics[width=0.4\textwidth]{GAN.png}
%     \caption{Feedback loop to solve Carlini \& Wagner optimisation problem}
%     \label{fig:CW_feedback loop gan}
% \end{figure}

Figure \ref{fig:proc} visualizes our framework and the end-to-end optimization. The generation of an adversarial image starts with the input of the target image. The framework internally initializes a latent object vector $z$ by a random draw from $\mathcal{N}(0, I)$ and overlays the picture generated by the GAN on the input image. The object detector scores each candidate bounding box we need for our loss function. The latent object representation $z$ is then updated using a suitable gradient descent method with the overall end-to-end gradient.

%in the direction indicated by the gradient such that the overall total loss (\ref{eq:opt}) is reduced.% Optimization is implemented according to gradient descent with the end-to-end gradient $z' = z - \lambda \nabla_z L(z)$.

Our proposed framework uses only the generator network of the GAN. We therefore just expect a (differentiable) function $\operatorname{GAN}(z)$ creating images form latent representations $z$. Such functions can potentially also be learned by other frameworks, e.g. (Variational) Auto Encoder \cite{kingma2013auto}. The only reason to use a GAN in our work is the typically more convincing perceptual quality of generated images.

\subsection{Expectation over Transformation}

To improve the robustness of the adversarial objects, we implemented the Expectation over Transformation method \cite{athalye2017synthesizing} using translation, brightness, size and smoothing transforms $\mathcal{T}$. We optimize the expectation value of \autoref{eq:opt} under application of such transforms to the generated object before patching them onto the background image. Integrating over transforms increased the success rate of adversarial object generation.

\subsection{Implementation}

\begin{algorithm}
\SetAlgoLined
\SetKwInOut{Input}{input}\SetKwInOut{Output}{output}
\Input{Background Image $x'$, Target Detection Class $c$, Generative Adversarial Network $\operatorname{GAN}$, Object Detector $D$, Learning Rate $\lambda$}
\Output{vector $z$}
$z \longleftarrow \operatorname{random}(\mathcal{N}(0,I))$\;
$counter \longleftarrow 0$\;
\While{$counter < 5$}{
$t \longleftarrow \operatorname{random}(\mathcal{T})$\;
$z \longleftarrow \operatorname{adamUpdate}(z, \nabla_z \mathcal{L}(z, c, t), \lambda)$\;
\eIf{$D(x' \, \oplus \, \operatorname{GAN}(z),\,c) \geq 0.95$}{
$counter$ ++\;
}{
$counter \longleftarrow 0$\;
}
}
\caption{Generating Unrestricted False Positive Adversarial Objects}
\label{alg:unr}
\end{algorithm}

We implement the method in TensorFlow 1.12 \cite{tensorflow_2020}. We import two frozen graphs into our framework. One corresponding to the target object detector and the other one corresponding to the GAN network. We have a complete view of these two graphs, and we are able to augment them such that we can skip nodes which loose gradient such as rounding. We optimize using TensorFlow's Adam implementation. In each iteration we apply a random transform to the generated image before stitching onto the background image. Because detection model is DNN we approximate $\mathcal{L}_0$ (\autoref{eq:l0}) with second to last layer values, before sigmoid function is applied. Loss function optimization is noisy. Therefore, optimization is terminated with success if the detection confidence for our target class surpasses 0.9 five iterations in a row within at most 2,000 iterations, otherwise the case fails. Algorithm \ref{alg:unr} shows the generation process.

\section{Evaluation}\label{sec:eval}

\subsection{Setup}

We illustrate our new attack on traffic sign object detection networks. The correct functionality of these detectors is important for autonomous driving applications. We attack three different pre-trained models: two Faster R-CNNs object detectors (using ResNet-101 and Inception v2 as CNN layer) and SSD Mobilenet v1. These models recognize three categories of traffic signs: prohibitory, mandatory and danger.
These networks were evaluated in the study on deep neural networks for traffic signs in \cite{arcos2018evaluation}. We chose these networks to cover various possible use cases: the Faster R-CNN networks have a high mean average precision (mAP), while the SSD Mobilenet has a fast inference time. Table \ref{tab:perf} summarizes the object detectors used in our experiments, and their performance, as reported in \cite{arcos2018evaluation}. In the rest of the evaluation section, we use the terms ResNet-101, Inception v2 and SSD Mobilenet to denote these three object detectors.

\begin{table}
\caption{Performance and execution time of object detectors \cite{arcos2018evaluation} used in evaluation.}\label{tab:perf}
\begin{center}
 \begin{tabular}{l  r  r }
 \toprule
 Object Detector & mAP & Inference [ms]\\
 \midrule
 ResNet-101 & 95.08 & 123 \\
 \midrule
 Inception v2& 90.62 & 59 \\
 \midrule
 SSD Mobilenet& 61.64 & 15 \\
 \bottomrule
\end{tabular}
\end{center}
\end{table}

\begin{figure*}
    \centering
    % flatex input: [traffic_signs/signs.tikz]
\begin{tikzpicture}
    
    \draw[rounded corners=5pt, line width = 4pt, color=red!30]
  (-0.75,0.75) rectangle (3.25,-5.25);
  
    \draw[rounded corners=5pt, line width = 4pt, color=blue!30]
  (3.5,0.75) rectangle (7.5,-5.25);
  
    \draw[rounded corners=5pt, line width = 4pt, color=black!30]
  (7.75,0.75) rectangle (11.75,-5.25);
  
  \node at (1.25, -4.75) {Prohibitory Signs};
  \node at (5.5, -4.75) {Mandatory Signs};
  \node at (9.75, -4.75) {Danger Signs};
  
  \node [anchor=east] at (-1, 0) {Traffic Signs};
  \node [anchor=east] at (-1, -1.25) {ResNet-101};
  \node [anchor=east] at (-1, -2.5) {Inception v2};
  \node [anchor=east] at (-1, -3.75) {SSD Mobilenet};
    
    \savebox{\picbox}{\includegraphics[width=1cm]{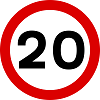}}
    \node (pro1) [minimum width=\wd\picbox,
        minimum height=\ht\picbox, path picture={
         \node at (path picture bounding box.center) {
            \usebox{\picbox}};
        }] at (0, 0) {};
        
    \savebox{\picbox}{\includegraphics[width=1cm]{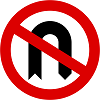}}
    \node (pro2) [minimum width=\wd\picbox,
        minimum height=\ht\picbox, path picture={
         \node at (path picture bounding box.center) {
            \usebox{\picbox}};
        }]  at (1.25, 0) {};
    
    \savebox{\picbox}{\includegraphics[width=1cm]{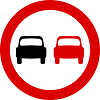}}
    \node (pro3) [minimum width=\wd\picbox,
        minimum height=\ht\picbox, path picture={
         \node at (path picture bounding box.center) {
            \usebox{\picbox}};
        }]at (2.5, 0) {};

    \savebox{\picbox}{\includegraphics[width=1cm]{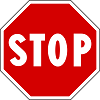}}
    \node (man1) [minimum width=\wd\picbox,
        minimum height=\ht\picbox, path picture={
         \node at (path picture bounding box.center) {
            \usebox{\picbox}};
        }]at (4.25, 0) {};
        
    \savebox{\picbox}{\includegraphics[width=1cm]{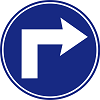}}
    \node (man2) [minimum width=\wd\picbox,
        minimum height=\ht\picbox, path picture={
         \node at (path picture bounding box.center) {
            \usebox{\picbox}};
        }]at (5.5, 0) {};
        
    \savebox{\picbox}{\includegraphics[width=1cm]{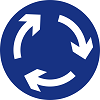}}
    \node (man3) [minimum width=\wd\picbox,
        minimum height=\ht\picbox, path picture={
         \node at (path picture bounding box.center) {
            \usebox{\picbox}};
        }]at (6.75, 0) {};
        
    \savebox{\picbox}{\includegraphics[width=1cm]{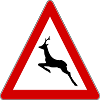}}
    \node (dan1) [minimum width=\wd\picbox,
        minimum height=\ht\picbox, path picture={
         \node at (path picture bounding box.center) {
            \usebox{\picbox}};
        }]at (8.5, 0) {};
        
    \savebox{\picbox}{\includegraphics[width=1cm]{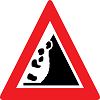}}
    \node (dan2) [minimum width=\wd\picbox,
        minimum height=\ht\picbox, path picture={
         \node at (path picture bounding box.center) {
            \usebox{\picbox}};
        }]at (9.75, 0) {};
        
    \savebox{\picbox}{\includegraphics[width=1cm]{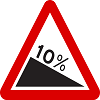}}
    \node (dan3) [minimum width=\wd\picbox,
        minimum height=\ht\picbox, path picture={
         \node at (path picture bounding box.center) {
            \usebox{\picbox}};
        }]at (11, 0) {};

    %%%%%%%%%%%%%%%%%%%%%%%%%%%%%%%%%%%%%%%%%%%%%%%%%%%%%%%%%%%%%%%%%%%%

    \savebox{\picbox}{\includegraphics[width=1cm]{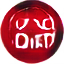}}
    \node (pro1) [minimum width=\wd\picbox,
        minimum height=\ht\picbox, path picture={
         \node at (path picture bounding box.center) {
            \usebox{\picbox}};
        }] at (0, -1.25) {};
        
    \savebox{\picbox}{\includegraphics[width=1cm]{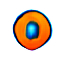}}
    \node (pro2) [minimum width=\wd\picbox,
        minimum height=\ht\picbox, path picture={
         \node at (path picture bounding box.center) {
            \usebox{\picbox}};
        }]  at (1.25, -1.25) {};
    
    \savebox{\picbox}{\includegraphics[width=1cm]{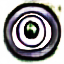}}
    \node (pro3) [minimum width=\wd\picbox,
        minimum height=\ht\picbox, path picture={
         \node at (path picture bounding box.center) {
            \usebox{\picbox}};
        }]at (2.5, -1.25) {};
    
    \savebox{\picbox}{\includegraphics[width=1cm]{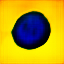}}
    \node (man1) [minimum width=\wd\picbox,
        minimum height=\ht\picbox, path picture={
         \node at (path picture bounding box.center) {
            \usebox{\picbox}};
        }]at (4.25, -1.25) {};
        
    \savebox{\picbox}{\includegraphics[width=1cm]{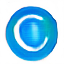}}
    \node (man2) [minimum width=\wd\picbox,
        minimum height=\ht\picbox, path picture={
         \node at (path picture bounding box.center) {
            \usebox{\picbox}};
        }]at (5.5, -1.25) {};
        
    \savebox{\picbox}{\includegraphics[width=1cm]{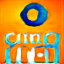}}
    \node (man3) [minimum width=\wd\picbox,
        minimum height=\ht\picbox, path picture={
         \node at (path picture bounding box.center) {
            \usebox{\picbox}};
        }]at (6.75, -1.25) {};

    \savebox{\picbox}{\includegraphics[width=1cm]{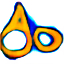}}
    \node (dan1) [minimum width=\wd\picbox,
        minimum height=\ht\picbox, path picture={
         \node at (path picture bounding box.center) {
            \usebox{\picbox}};
        }]at (8.5, -1.25) {};
        
    \savebox{\picbox}{\includegraphics[width=1cm]{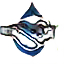}}
    \node (dan2) [minimum width=\wd\picbox,
        minimum height=\ht\picbox, path picture={
         \node at (path picture bounding box.center) {
            \usebox{\picbox}};
        }]at (9.75, -1.25) {};
        
    \savebox{\picbox}{\includegraphics[width=1cm]{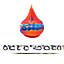}}
    \node (dan3) [minimum width=\wd\picbox,
        minimum height=\ht\picbox, path picture={
         \node at (path picture bounding box.center) {
            \usebox{\picbox}};
        }]at (11, -1.25) {};

    %%%%%%%%%%%%%%%%%%%%%%%%%%%%%%%%%%%%%%%%%%%%%%%%%%%%%%%%%%%%%%%%%%%%

    \savebox{\picbox}{\includegraphics[width=1cm]{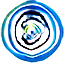}}
    \node (pro1) [minimum width=\wd\picbox,
        minimum height=\ht\picbox, path picture={
         \node at (path picture bounding box.center) {
            \usebox{\picbox}};
        }] at (0, -2.5) {};
        
    \savebox{\picbox}{\includegraphics[width=1cm]{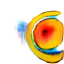}}
    \node (pro2) [minimum width=\wd\picbox,
        minimum height=\ht\picbox, path picture={
         \node at (path picture bounding box.center) {
            \usebox{\picbox}};
        }]  at (1.25, -2.5) {};
    
    \savebox{\picbox}{\includegraphics[width=1cm]{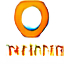}}
    \node (pro3) [minimum width=\wd\picbox,
        minimum height=\ht\picbox, path picture={
         \node at (path picture bounding box.center) {
            \usebox{\picbox}};
        }]at (2.5, -2.5) {};

    \savebox{\picbox}{\includegraphics[width=1cm]{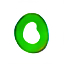}}
    \node (man1) [minimum width=\wd\picbox,
        minimum height=\ht\picbox, path picture={
         \node at (path picture bounding box.center) {
            \usebox{\picbox}};
        }]at (4.25, -2.5) {};
        
    \savebox{\picbox}{\includegraphics[width=1cm]{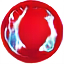}}
    \node (man2) [minimum width=\wd\picbox,
        minimum height=\ht\picbox, path picture={
         \node at (path picture bounding box.center) {
            \usebox{\picbox}};
        }]at (5.5, -2.5) {};
        
    \savebox{\picbox}{\includegraphics[width=1cm]{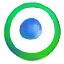}}
    \node (man3) [minimum width=\wd\picbox,
        minimum height=\ht\picbox, path picture={
         \node at (path picture bounding box.center) {
            \usebox{\picbox}};
        }]at (6.75, -2.5) {};
        
    \savebox{\picbox}{\includegraphics[width=1cm]{v2/kl_3_smoothin_10.jpg}}
    \node (dan1) [minimum width=\wd\picbox,
        minimum height=\ht\picbox, path picture={
         \node at (path picture bounding box.center) {
            \usebox{\picbox}};
        }]at (8.5, -2.5) {};
        
    \savebox{\picbox}{\includegraphics[width=1cm]{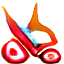}}
    \node (dan2) [minimum width=\wd\picbox,
        minimum height=\ht\picbox, path picture={
         \node at (path picture bounding box.center) {
            \usebox{\picbox}};
        }]at (9.75, -2.5) {};
        
    \savebox{\picbox}{\includegraphics[width=1cm]{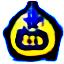}}
    \node (dan3) [minimum width=\wd\picbox,
        minimum height=\ht\picbox, path picture={
         \node at (path picture bounding box.center) {
            \usebox{\picbox}};
        }]at (11, -2.5) {};

    %%%%%%%%%%%%%%%%%%%%%%%%%%%%%%%%%%%%%%%%%%%%%%%%%%%%%%%%%%%%%%%%%%%%

    \savebox{\picbox}{\includegraphics[width=1cm]{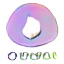}}
    \node (pro1) [minimum width=\wd\picbox,
        minimum height=\ht\picbox, path picture={
         \node at (path picture bounding box.center) {
            \usebox{\picbox}};
        }] at (0, -3.75) {};
        
    \savebox{\picbox}{\includegraphics[width=1cm]{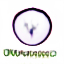}}
    \node (pro2) [minimum width=\wd\picbox,
        minimum height=\ht\picbox, path picture={
         \node at (path picture bounding box.center) {
            \usebox{\picbox}};
        }]  at (1.25, -3.75) {};
    
    \savebox{\picbox}{\includegraphics[width=1cm]{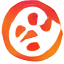}}
    \node (pro3) [minimum width=\wd\picbox,
        minimum height=\ht\picbox, path picture={
         \node at (path picture bounding box.center) {
            \usebox{\picbox}};
        }]at (2.5, -3.75) {};
        
    \savebox{\picbox}{\includegraphics[width=1cm]{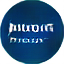}}
    \node (man1) [minimum width=\wd\picbox,
        minimum height=\ht\picbox, path picture={
         \node at (path picture bounding box.center) {
            \usebox{\picbox}};
        }]at (4.25, -3.75) {};
        
    \savebox{\picbox}{\includegraphics[width=1cm]{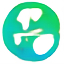}}
    \node (man2) [minimum width=\wd\picbox,
        minimum height=\ht\picbox, path picture={
         \node at (path picture bounding box.center) {
            \usebox{\picbox}};
        }]at (5.5, -3.75) {};
        
    \savebox{\picbox}{\includegraphics[width=1cm]{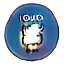}}
    \node (man3) [minimum width=\wd\picbox,
        minimum height=\ht\picbox, path picture={
         \node at (path picture bounding box.center) {
            \usebox{\picbox}};
        }]at (6.75, -3.75) {};
        
    \savebox{\picbox}{\includegraphics[width=1cm]{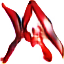}}
    \node (dan1) [minimum width=\wd\picbox,
        minimum height=\ht\picbox, path picture={
         \node at (path picture bounding box.center) {
            \usebox{\picbox}};
        }]at (8.5, -3.75) {};
        
    \savebox{\picbox}{\includegraphics[width=1cm]{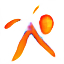}}
    \node (dan2) [minimum width=\wd\picbox,
        minimum height=\ht\picbox, path picture={
         \node at (path picture bounding box.center) {
            \usebox{\picbox}};
        }]at (9.75, -3.75) {};
        
    \savebox{\picbox}{\includegraphics[width=1cm]{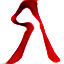}}
    \node (dan3) [minimum width=\wd\picbox,
        minimum height=\ht\picbox, path picture={
         \node at (path picture bounding box.center) {
            \usebox{\picbox}};
        }]at (11, -3.75) {};

    \end{tikzpicture}
% flatex input end: [traffic_signs/signs.tikz]

%$z \longleftarrow z - \lambda \nabla_z L(z, c, t)$
    \caption{Adversarial objects for three classes of traffic signs attacking three different object detection networks}
    \label{fig:res}
\end{figure*}

In our experiments, we use the logo generator iWGAN-LC \cite{sage2018logo} to synthesize the adversarial objects. For this logo GAN, the output is a square logo of size $64 \times 64$ pixels. An attacker could add the realistically looking logos in places visible to the camera used for the traffic sign detection (other cars in traffic, billboards). Figure \ref{fig:res} shows  sample adversarial logos generated by the logo GAN for each category of traffic signs, together with real traffic signs from each category of the object detectors.

To propagate the gradient through the Faster R-CNN neural networks, we used the algorithm presented in \cite{chen2018shapeshifter}, which overcomes the non-differentiability of the two-stage process of these networks.

Since SSD Mobilenet is a one phase detector, we adapt the generation process. As a first step, SSD Mobilenet resizes original image of size $1200 \times 720$ to $300 \times 300$. The logo is resized $4 \times$ on x-axis and $2.4\times$ on y-axis. The resizing operation is differentiable, therefore the gradient can be propagated through. Downsizing means that the gradient is distributed among multiple pixels of the original logo. The generation process stagnates and does not produce adversarial objects. To address this challenge, we exploit the detection process of SSD Mobilenet, which detects bounding boxes of various scales. We resize the background image to $300 \times 300$ and overlap the object over the resized image. While the adversarial object is optimized on large bounding boxes of SSD, it is also effective on small bounding boxes.

% flatex input: [logos.tex]
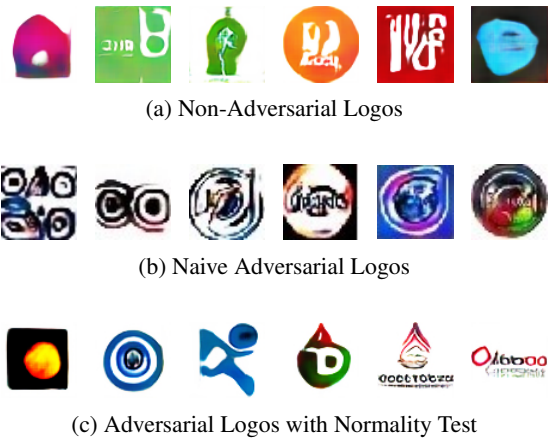
\begin{figure}
    \centering
    \begin{subfigure}[b]{0.5\textwidth}
    \centering
    \begin{tikzpicture}
    
    \savebox{\picbox}{\includegraphics[width=1cm]{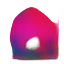}}
    \node (pro1) [minimum width=\wd\picbox,
        minimum height=\ht\picbox, path picture={
         \node at (path picture bounding box.center) {
            \usebox{\picbox}};
        }] at (0, 0) {};
    
    \savebox{\picbox}{\includegraphics[width=1cm]{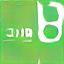}}
    \node (pro1) [minimum width=\wd\picbox,
        minimum height=\ht\picbox, path picture={
         \node at (path picture bounding box.center) {
            \usebox{\picbox}};
        }] at (1.25, 0) {};
        
    \savebox{\picbox}{\includegraphics[width=1cm]{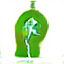}}
    \node (pro1) [minimum width=\wd\picbox,
        minimum height=\ht\picbox, path picture={
         \node at (path picture bounding box.center) {
            \usebox{\picbox}};
        }] at (2.5, 0) {};
        
    \savebox{\picbox}{\includegraphics[width=1cm]{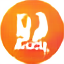}}
    \node (pro1) [minimum width=\wd\picbox,
        minimum height=\ht\picbox, path picture={
         \node at (path picture bounding box.center) {
            \usebox{\picbox}};
        }] at (3.75, 0) {};
        
    \savebox{\picbox}{\includegraphics[width=1cm]{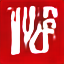}}
    \node (pro1) [minimum width=\wd\picbox,
        minimum height=\ht\picbox, path picture={
         \node at (path picture bounding box.center) {
            \usebox{\picbox}};
        }] at (5, 0) {};
        
    \savebox{\picbox}{\includegraphics[width=1cm]{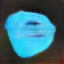}}
    \node (pro1) [minimum width=\wd\picbox,
        minimum height=\ht\picbox, path picture={
         \node at (path picture bounding box.center) {
            \usebox{\picbox}};
        }] at (6.25, 0) {};
    
    \end{tikzpicture}
    \caption{Non-Adversarial Logos}
    \end{subfigure}
    
    \begin{subfigure}[b]{0.5\textwidth}
    \vspace{0.5cm}
    \centering
    \begin{tikzpicture}
    
    \savebox{\picbox}{\includegraphics[width=1cm]{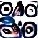}}
    \node (pro1) [minimum width=\wd\picbox,
        minimum height=\ht\picbox, path picture={
         \node at (path picture bounding box.center) {
            \usebox{\picbox}};
        }] at (0, 0) {};
    
    \savebox{\picbox}{\includegraphics[width=1cm]{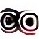}}
    \node (pro1) [minimum width=\wd\picbox,
        minimum height=\ht\picbox, path picture={
         \node at (path picture bounding box.center) {
            \usebox{\picbox}};
        }] at (1.25, 0) {};
        
    \savebox{\picbox}{\includegraphics[width=1cm]{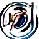}}
    \node (pro1) [minimum width=\wd\picbox,
        minimum height=\ht\picbox, path picture={
         \node at (path picture bounding box.center) {
            \usebox{\picbox}};
        }] at (2.5, 0) {};
        
    \savebox{\picbox}{\includegraphics[width=1cm]{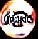}}
    \node (pro1) [minimum width=\wd\picbox,
        minimum height=\ht\picbox, path picture={
         \node at (path picture bounding box.center) {
            \usebox{\picbox}};
        }] at (3.75, 0) {};
        
    \savebox{\picbox}{\includegraphics[width=1cm]{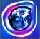}}
    \node (pro1) [minimum width=\wd\picbox,
        minimum height=\ht\picbox, path picture={
         \node at (path picture bounding box.center) {
            \usebox{\picbox}};
        }] at (5, 0) {};
        
    \savebox{\picbox}{\includegraphics[width=1cm]{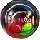}}
    \node (pro1) [minimum width=\wd\picbox,
        minimum height=\ht\picbox, path picture={
         \node at (path picture bounding box.center) {
            \usebox{\picbox}};
        }] at (6.25, 0) {};
    
    \end{tikzpicture}
    \caption{Naive Adversarial Logos}
    \end{subfigure}
    
    \begin{subfigure}[b]{0.5\textwidth}
    \vspace{0.5cm}
    \centering
    \begin{tikzpicture}
    
    \savebox{\picbox}{\includegraphics[width=1cm]{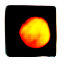}}
    \node (pro1) [minimum width=\wd\picbox,
        minimum height=\ht\picbox, path picture={
         \node at (path picture bounding box.center) {
            \usebox{\picbox}};
        }] at (0, 0) {};
    
    \savebox{\picbox}{\includegraphics[width=1cm]{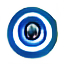}}
    \node (pro1) [minimum width=\wd\picbox,
        minimum height=\ht\picbox, path picture={
         \node at (path picture bounding box.center) {
            \usebox{\picbox}};
        }] at (1.25, 0) {};
        
    \savebox{\picbox}{\includegraphics[width=1cm]{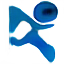}}
    \node (pro1) [minimum width=\wd\picbox,
        minimum height=\ht\picbox, path picture={
         \node at (path picture bounding box.center) {
            \usebox{\picbox}};
        }] at (2.5, 0) {};
        
    \savebox{\picbox}{\includegraphics[width=1cm]{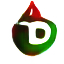}}
    \node (pro1) [minimum width=\wd\picbox,
        minimum height=\ht\picbox, path picture={
         \node at (path picture bounding box.center) {
            \usebox{\picbox}};
        }] at (3.75, 0) {};
        
    \savebox{\picbox}{\includegraphics[width=1cm]{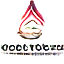}}
    \node (pro1) [minimum width=\wd\picbox,
        minimum height=\ht\picbox, path picture={
         \node at (path picture bounding box.center) {
            \usebox{\picbox}};
        }] at (5, 0) {};
        
    \savebox{\picbox}{\includegraphics[width=1cm]{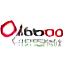}}
    \node (pro1) [minimum width=\wd\picbox,
        minimum height=\ht\picbox, path picture={
         \node at (path picture bounding box.center) {
            \usebox{\picbox}};
        }] at (6.25, 0) {};
    
    \end{tikzpicture}
    \caption{Adversarial Logos with Normality Test}
    \end{subfigure}
    
    \caption{Comparison between non-adversarial, naive adversarial, and adversarial logos generated by adding the normality test to the loss function.}
    \label{fig:logos}
\end{figure}
% flatex input end: [logos.tex]

%$z \longleftarrow z - \lambda \nabla_z L(z, c, t)$

\subsection{Normality Test in Loss Function}

% \begin{figure}
%     \centering
%     \includegraphics{test_traffic_101_no_norm_high_var_crop.jpg}
%     \caption{Naively generated adversarial stickers with high variance.}
%     \label{fig:naive}
% \end{figure}

To illustrate the importance of the normality test in the loss function, we generate three types of logos. First, we generate benign logos by randomly sampling a latent vector from normal distribution $\mathcal{N}(0,I)$.
Second, we generate adversarial image by minimizing detection loss only thereby ignoring the distribution loss by setting $\kappa=0$ in (\autoref{eq:opt}) (naive adversarial logos). The quality of naive adversarial logos is visibly lower. The propagated gradient can change the input latent vector such that the likelihood of a vector being sampled from normal distribution is low. The probability that these latent variables are encountered during training process of the GAN is negligible. Therefore, the generator outputs a logo which looks artificial.
The third group of logos are adversarial logos that were generated by using the normality test in the loss function (\autoref{eq:kl}).

\autoref{fig:logos} shows all the three types of logos. We observe that the logos with a high variance look unnatural, whereas the non-adversarial and adversarial logos have a better quality. Importantly, we observe that the adversarial logos have a similar natural look compared to the non-adversarial logos.

\subsection{Improving Adversarial Robustness}

To improve the robustness of the adversarial objects, we add random transformations to the input during the generation process, described in  \cite{athalye2017synthesizing}. Additionally, these transformations speed up the generation process, due to a broader exploration of the latent space. We implement four types of transformations:
% We wanted our adversarial stickers to be successful in physical world. Sticker should be able to fool the object detection network in various conditions the photo was taken in. In order to reduce complexity we simplified different conditions in four categories.
\begin{inparaenum}[1)]
    \item Translation: the adversarial object is shifted to different position onto the input image;
    \item Brightness: the overall brightness of image with adversarial object is changed;
    \item Size: the adversarial object is resized;
    \item Smoothing: a smoothing Bayesian filter is applied to the image with adversarial object.
\end{inparaenum}

At each iteration, we add small random perturbations to these parameters. We build three experiments, to simulate various physical world conditions. We see that the robustness of the adversarial logos with transformation is better, but the difference is small. However, the success rate of generating an adversarial logo in $2000$ iterations increased by $10\%$. Our hypothesis is that the logo GAN creates a robust image with large areas of the same color. Moreover, these areas are clearly separated. The robustness gains are small, but the success rate of generating adversarial logos is improved by the transformations.

When using the transformations, individual success rates of ResNet-101, Inception v2, and SSD Mobilenet are $67\%$, $76\%$ and $5\%$ respectively. Even though the success rate for SSD Mobilenet is low, attacker is not constrained by time to perform the attack.

\subsection{Transferable Adversarial Objects}

Given the three object detectors, we evaluate the transferability of the adversarial objects. Even if an attacker does not have access to the object detector architecture, it is possible to perform black-box attacks by training different DNN and generating adversarial object on it.

\begin{table}
\caption{Transferability of adversarial objects between different object detectors. The rows correspond to the originally attacked model. The columns show transferability to the target models.}\label{tab:trans}
\begin{center}
\begin{tabular}{ l r r r }
 \toprule
 & Resnet-101 & Inception v2 & SSD \\
 \midrule
 Resnet-101 & \makecell{--} & 64.36\% & 16.83\% \\
 \midrule
 Inception v2& 50.88\% & \makecell{--} & 28.07\% \\
 \midrule
 SSD & 21.43\% & 39.29\% & \makecell{--} \\
 \bottomrule

\end{tabular}
\end{center}
\end{table}

\autoref{tab:trans} shows our transferability experiment results. The two Faster R-CNNs have a high transferability, whereas the attacks are less transferable to the SSD Mobilenet. Given that the Faster R-CNNs share a significant portion of the DNN architecture, these results are expected. Overall, \autoref{tab:trans} shows that the adversarial objects generated with our algorithm are transferable between object detectors, even if they are based on completely different architectures.

\subsection{Physical world experiment}

For the physical test, we generate $9$ adversarial logos for all pairs of attacked classes (prohibitory, mandatory, danger) and detection networks (Faster R-CNN ResNet-101, Faster R-CNN Inception v2, SSD Mobilnet v1). After the generation, the logos are printed using a standard office printer on a regular paper, resized to $20\; \text{cm} \times 20 \; \text{cm}$.

\begin{figure}
\centering
\includegraphics[width=0.45\textwidth]{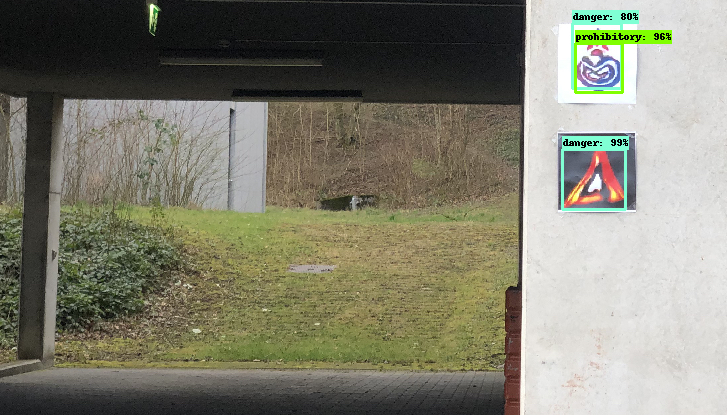}
\caption{Photo taken from position 4 with adversarial objects identified as traffic signs by the object detectors.}\label{fig:photo}
\end{figure}

\begin{figure}
\vspace{0.5cm}
\centering
% flatex input: [phys.tikz]
\begin{tikzpicture}
\tikzstyle{arrow} = [color=black!30, line width = 2pt,|-|]

\tikzstyle{arrow2} = [color=blue!20, line width = 1pt,dashed,-,>=angle 60]

\draw [arrow2] (5.5, 7) -- (7.5, 9.9);
\draw [arrow2] (6.5, 7) -- (7.5, 9.9);
\draw [arrow2] (7.5, 7) -- (7.5, 9.9);
\draw [arrow2] (5.5, 8.5) -- (7.5, 9.9);
%\draw [arrow2] (6.5, 8.5) -- (7.5, 9.9);
%\draw [arrow2] (7.5, 8.5) -- (7.5, 9.9);

\draw [color=red, line width = 4pt](7.2,10) -- (7.8,10);
\filldraw (7.5, 8.5) circle (3pt);
\filldraw (7.5, 7) circle (3pt);
\filldraw (6.5, 8.5) circle (3pt);
\filldraw (6.5, 7) circle (3pt);
\filldraw (5.5, 8.5) circle (3pt);
\filldraw (5.5, 7) circle (3pt);
\node [anchor=north east] at (7.5, 8.5) {1};
\node [anchor=north east] at (6.5, 8.5) {2};
\node [anchor=north east] at (5.5, 8.5) {3};
\node [anchor=north east] at (7.5, 7) {4};
\node [anchor=north east] at (6.5, 7) {5};
\node [anchor=north east] at (5.5, 7) {6};
\node [anchor=south] at (7.5, 10) {Adversarial Object};
\node [anchor=west, xshift=3pt] at (8.1, 8.5) {5m};
%\node [anchor=west, xshift=3pt] at (7.5, 8.5) {2.5m};
\node [anchor=south] at (6.5, 8.9) {4m};

\draw [arrow] (8.1, 10.05) -- (8.1, 6.95);
\draw [arrow] (7.55, 8.9) -- (5.45, 8.9);

\end{tikzpicture}
% flatex input end: [phys.tikz]

% & 39.29\% & \makecell{--} \\
\caption{Bird's-eye view of the six positions of the camera for the physical world experiment.}\label{fig:phys}
\end{figure}
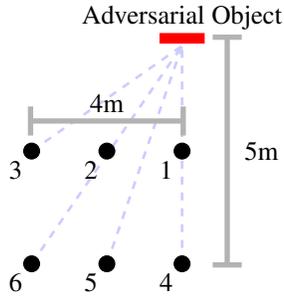

For each logo, we take six photos from different distances and angles. \autoref{fig:phys} shows the six positions we have used. Photos are taken using a iPhone X camera with telephoto f/2.4 aperture. \autoref{fig:photo} shows an example photo taken from position 4.

Finally, the photos are down-sampled to $1200 \times 720$ pixels, and fed into the three detection networks. \autoref{tab:phys} shows the confidence of each network in detecting given class for each adversarial logo. Positions $3$ and $6$ give the lowest confidence values on average, due to the angle and distance from the logo. We observe the transferability of adversarial objects in physical world between the two Faster R-CNNs as well. SSD Mobilet under-performs due to its low detection performance. Additionally, we test two non-adversarial logos in the physical world and they are not detected from any of the six positions.

% flatex input: [phy-table.tex]
\begin{table*}
\caption{Detection confidence of object detection DNNs in physical world experiment for 9 adversarial objects.}
\label{tab:phys}
\centering
\small
\begin{tabular}{ c c c c c c c c c c }
 \toprule
 \makecell{ResNet-101 \\ Inception v2\\ SSD }
 & \makecell{\includegraphics[width=1cm]{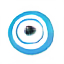}} & \makecell{\includegraphics[width=1cm]{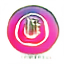}} &
 \makecell{\includegraphics[width=1cm]{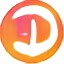}} &
 \makecell{\includegraphics[width=1cm]{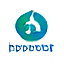}} & \makecell{\includegraphics[width=1cm]{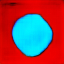}} &
 \makecell{\includegraphics[width=1cm]{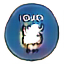}} &
 \makecell{\includegraphics[width=1cm]{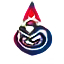}}
 & \makecell{\includegraphics[width=1cm]{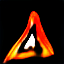}} &
 \makecell{\includegraphics[width=1cm]{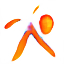}} \\ 
 \midrule
 Category & \multicolumn{3}{|c}{Prohibitory} & \multicolumn{3}{|c}{Mandatory} & \multicolumn{3}{|c}{Danger} \\
 \midrule
  Target DNN & ResNet & Inception & SSD & ResNet & Inception & SSD & ResNet & Inception & SSD \\
 \midrule
 Pos. 1 & \makecell{99\% \\ 98\% \\ --}& \makecell{98\% \\ 99\% \\ 99\%}& \makecell{--\\ 54\%\\ 32\%} & \makecell{99\% \\ 99\% \\ --}& \makecell{99\% \\ 99\% \\ --}& \makecell{--\\ -- \\ 96\%} & \makecell{99\% \\ 98\% \\ --}&  \makecell{-- \\ 94\% \\ --}& \makecell{--\\ --\\ --} \\ 
 \midrule
 Pos. 2 & \makecell{ 99\% \\ 99\% \\ --} & \makecell{ 98\% \\ 99\% \\ 67\%} & \makecell{60\% \\ 39\%\\ --} & \makecell{ 99\% \\ 99\% \\ --} & \makecell{ 99\% \\ 99\% \\ --} & \makecell{--\\ 76\%\\ --} & \makecell{ 99\% \\ 99\% \\ --} & \makecell{ 99\% \\ 99\% \\ 43\%} & \makecell{--\\ 41\% \\ 44\%} \\ 
 \midrule
 Pos. 3 & \makecell{ 91\% \\ 92\% \\ --} & \makecell{ 95\% \\ 98\% \\ --} & \makecell{--\\ 36\%\\ --} & \makecell{ 99\% \\ 76\% \\ --} & \makecell{ --\\ 99\% \\ --}& \makecell{--\\ 93\%\\ --} & \makecell{ --\\ 20\% \\ --} &  \makecell{ 99\% \\ 99\% \\ 10\%} & \makecell{--\\ --\\ 24\%}  \\ 
 \midrule
 Pos. 4 & \makecell{ 99\% \\ 99\% \\ --} & \makecell{ 99\% \\ 99\% \\ 99\%} & \makecell{91\%\\ 35\%\\ 97\%} & \makecell{ 99\% \\ 99\% \\ --} & \makecell{ 99\% \\ 99\% \\ --} & \makecell{ 99\%\\ 97\% \\ 89\%} & \makecell{ 99\% \\ 80\% \\ --} & \makecell{ 99\% \\ 99\% \\ 69\%} & \makecell{--\\ 95\% \\ --}\\ 
 \midrule
 Pos. 5 & \makecell{ 97\% \\ 99\% \\ --} & \makecell{ 99\% \\ 99\% \\ 91\%} & \makecell{54\%\\ 45\% \\ --} & \makecell{ 99\% \\ 99\% \\ --} & \makecell{ 99\% \\ 99\% \\ --} & \makecell{99\%\\ 98\% \\ 99\%} & \makecell{ 99\% \\ 89\% \\ --} & \makecell{ 99\% \\ 99\% \\ 79\% } & \makecell{--\\ 57\%\\ 99\%} \\ 
 \midrule
 Pos. 6 & \makecell{ 84\% \\ 99\% \\ --} & \makecell{ 99\% \\ 99\% \\ --} & \makecell{89\%\\ 96\% \\ --} & \makecell{ 99\% \\ -- \\ --} & \makecell{ 31\% \\ 90\% \\ --}  & \makecell{99\%\\ 99\%\\ --} & \makecell{ -- \\11\% \\ --}& \makecell{99\% \\ 99\% \\ --} & \makecell{-- \\ --\\ -- } \\ 
 \bottomrule
\end{tabular}
\end{table*}
% flatex input end: [phy-table.tex]

% & 39.29\% & \makecell{--} \\

\subsection{CIFAR-10 SNGAN}

% flatex input: [cifar-10.tex]
\begin{figure}
    \centering
    \begin{subfigure}[b]{0.5\textwidth}
    \centering
    \begin{tikzpicture}
    
    \savebox{\picbox}{\includegraphics[width=1cm]{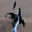}}
    \node (pro1) [minimum width=\wd\picbox,
        minimum height=\ht\picbox, path picture={
         \node at (path picture bounding box.center) {
            \usebox{\picbox}};
        }] at (0, 0) {};
    
    \savebox{\picbox}{\includegraphics[width=1cm]{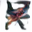}}
    \node (pro1) [minimum width=\wd\picbox,
        minimum height=\ht\picbox, path picture={
         \node at (path picture bounding box.center) {
            \usebox{\picbox}};
        }] at (1.25, 0) {};
        
    \savebox{\picbox}{\includegraphics[width=1cm]{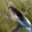}}
    \node (pro1) [minimum width=\wd\picbox,
        minimum height=\ht\picbox, path picture={
         \node at (path picture bounding box.center) {
            \usebox{\picbox}};
        }] at (2.5, 0) {};
        
    \savebox{\picbox}{\includegraphics[width=1cm]{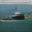}}
    \node (pro1) [minimum width=\wd\picbox,
        minimum height=\ht\picbox, path picture={
         \node at (path picture bounding box.center) {
            \usebox{\picbox}};
        }] at (3.75, 0) {};
        
    \savebox{\picbox}{\includegraphics[width=1cm]{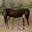}}
    \node (pro1) [minimum width=\wd\picbox,
        minimum height=\ht\picbox, path picture={
         \node at (path picture bounding box.center) {
            \usebox{\picbox}};
        }] at (5, 0) {};
        
    \savebox{\picbox}{\includegraphics[width=1cm]{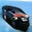}}
    \node (pro1) [minimum width=\wd\picbox,
        minimum height=\ht\picbox, path picture={
         \node at (path picture bounding box.center) {
            \usebox{\picbox}};
        }] at (6.25, 0) {};
    
    \end{tikzpicture}
    \caption{Non-adversarial Objects}
    \end{subfigure}
    \begin{subfigure}[b]{0.5\textwidth}
    \vspace{0.5cm}
    \centering
    \begin{tikzpicture}
    
    \savebox{\picbox}{\includegraphics[width=1cm]{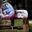}}
    \node (pro1) [minimum width=\wd\picbox,
        minimum height=\ht\picbox, path picture={
         \node at (path picture bounding box.center) {
            \usebox{\picbox}};
        }] at (0, 0) {};
    
    \savebox{\picbox}{\includegraphics[width=1cm]{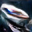}}
    \node (pro1) [minimum width=\wd\picbox,
        minimum height=\ht\picbox, path picture={
         \node at (path picture bounding box.center) {
            \usebox{\picbox}};
        }] at (1.25, 0) {};
        
    \savebox{\picbox}{\includegraphics[width=1cm]{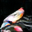}}
    \node (pro1) [minimum width=\wd\picbox,
        minimum height=\ht\picbox, path picture={
         \node at (path picture bounding box.center) {
            \usebox{\picbox}};
        }] at (2.5, 0) {};
        
    \savebox{\picbox}{\includegraphics[width=1cm]{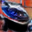}}
    \node (pro1) [minimum width=\wd\picbox,
        minimum height=\ht\picbox, path picture={
         \node at (path picture bounding box.center) {
            \usebox{\picbox}};
        }] at (3.75, 0) {};
        
    \savebox{\picbox}{\includegraphics[width=1cm]{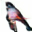}}
    \node (pro1) [minimum width=\wd\picbox,
        minimum height=\ht\picbox, path picture={
         \node at (path picture bounding box.center) {
            \usebox{\picbox}};
        }] at (5, 0) {};
        
    \savebox{\picbox}{\includegraphics[width=1cm]{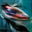}}
    \node (pro1) [minimum width=\wd\picbox,
        minimum height=\ht\picbox, path picture={
         \node at (path picture bounding box.center) {
            \usebox{\picbox}};
        }] at (6.25, 0) {};
    
    \end{tikzpicture}
    \caption{Adversarial Objects}
    \end{subfigure}
    \caption{Comparison between non-adversarial and adversarial objects generated by the SNGAN trained on the CIFAR-10 dataset.}
    \label{fig:sngan}
\end{figure}
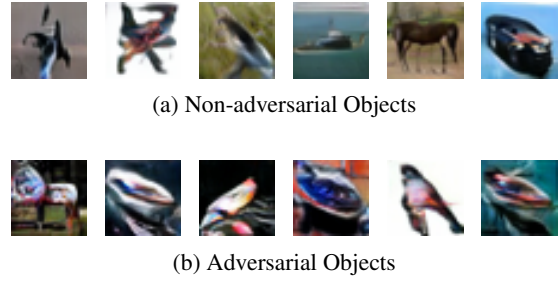
% flatex input end: [cifar-10.tex]

% & 39.29\% & \makecell{--} \\

To further evaluate our attack, we train a SNGAN \cite{miyato2018spectral} using the CIFAR-10 dataset. We have used the implementation described in \cite{yang}. \autoref{fig:sngan} shows samples of adversarial objects generated by this GAN, compared to non-adversarial objects. We observe that the non-adversarial examples have a similar quality with the adversarial objects. This strengthens our confidence that our attack method generates adversarial objects that look as natural as the normal images synthesized by the generative model.

\section{RELATED WORK} \label{sec:related}

\cite{song2018unconstrained} employs GAN networks to create examples which a human observer classifies differently than a classifier. The main difference from our work is the attack setting. In their work, the GAN creates images from the same domain as the target classifier. We can use off-the-shelf pre-trained GANs that synthesize images from domains unrelated to the detected classes of the object detectors (e.g., if the object detector identifies animals, our approach can use a GAN that creates cars as adversarial objects). Our attack can be easier to perform in the physical world and more covert.

\cite{wang2019nonconstrained, xiao2018generating} use GANs to directly create adversarial examples without an iterative method. \cite{wang2019nonconstrained} creates unrestricted examples, whereas \cite{xiao2018generating} uses GANs to create perturbation-based adversarial examples. \cite{zhao2017generating} uses a GAN and an inverter to find the latent variables. Their attack works by finding a good region of the latent space which creates adversarial examples through a random search algorithm, unlike our approach based on gradient descent. Their work focuses on black-box attacks. \cite{wang2019nonconstrained, xiao2018generating, zhao2017generating} need a specially-constructed GAN, whereas our method can work with pre-trained GANs.

All of the previously mentioned papers explore attacks against classifiers. In our work we explore attacks against object detectors. Additionally, none of them explores attacks in physical world. We believe that real-world perturbations represent a high, as these attacks do not require the access to the camera.

%Conversely to them we explored using already existing GAN trained for different task. This shows how an unskilled attacker can still perform this attack, without a knowledge of GAN networks.

% \cite{sitawarin2018darts} have a similar use case as our work. They try to sell logos as a poisonous traffic signs, such that a classifier finds a traffic sign in them. They initialize the optimisation with a existing logo and try to find minimal perturbations to it. They use $l_p$ distance for finding minimal perturbations.

% Looking at examples from papers, we believe they are suspicious, and they are defensible by adversarial training. To a human eye, security personal, and or law enforcement they would be easily identifiable. Low quality camera might not pick the features of adversarial sticker such as Figure \ref{fig:fp}. We address both of these issues in our work by using GAN network which produces both robust images and assimilates in a use case setting.

In recent work, \cite{kurakin2016adversarial, sharif2016accessorize, eykholt2018classification, eykholt2018detection, chen2018shapeshifter} explored adversarial examples in physical world. They work by creating adversarial stickers/patches which are put onto other objects, e.g., stop signs. After the camera takes a picture, the augmented stop sign is misclassified/undetected. Moreover, \cite{eykholt2018detection} looked at creating false positive attacks using current techniques for creating perturbation-based adversarial examples. The detection model recognized the adversarial object as a stop sign, even thought to a human observer it does not look like one. The potential limitation of these approaches is the high salience of such objects. We believe that if the attacker launches a physical attack, it is easy to identify the malicious object and indict the attacker. In our work, the GAN network produces robust adversarial objects which are assimilated in the real world scenery.

The detectors used in \cite{eykholt2018detection, chen2018shapeshifter, sitawarin2018darts} are not accessible. The missing performance statistics make it hard to judge the attack's strength. In our work, the detectors are chosen from a published work in traffic sign detection \cite{arcos2018evaluation}.

% \subsection{Defensive Techniques}

% We believe our work is also robust under various defensive techniques.

% Adversarial Training \cite{madry2017towards} would not be effective due to different domain the pictures are coming from. \cite{song2018unconstrained} already showed that unrestricted adversarial examples are not prevented with adversarial training. Uncorrelated adversarial objects would not be seen during adversarial training. Attacker can always switch to different GAN if adversarial training is successful.

% Adversarial Detection again would not be able to know what to expect, therefore training wouldn't be considering.

% Input Reconstruction

% The only prevention is robust training, which might be very difficult, or in terms of SSD Mobilenet, less performant detectors might not be successible to this attack.
% \todo{write something clever here}

\section{Conclusion}
We defined a new attack for object detectors, based on unrestricted adversarial examples. To synthesize adversarial objects, we use off-the-shelf pre-trained GANs that generate images unrelated to the detected classes of the target neural network. Importantly, the synthesized adversarial objects are indistinguishable from non-adversarial outputs of the GANs, from a human's perspective. We experimented with the Faster  R-CNN ResNet-101, Inception v2 and SSD Mobilenet object detectors trained for traffic sign recognition and used two pre-trained generative models, SNGAN trained on CIFAR-10 and logo generating iWGAN-LC. The evaluation results show that the adversarial objects are transferable between target neural networks (between $16\%$ and $64\%$). Moreover, we validated the adversarial objects in a physical world setup, which demonstrated their robustness to camera angle and distance to the camera. By introducing unrestricted false positive adversarial objects, we extend the space of attacks and open new directions to investigate the reliability of object detectors.

\small

%*flatex input: [main.bbl]

% flatex input end: [main.bbl]
%FLATEX-REM:\bibliographystyle{apalike}
%FLATEX-REM:{\small \bibliography{bibl.bib}}

\begin{thebibliography}{}

\small

\bibitem[Arcos-Garcia et~al., 2018]{arcos2018evaluation}
Arcos-Garcia, A., Alvarez-Garcia, J.~A., and Soria-Morillo, L.~M. (2018).
\newblock Evaluation of deep neural networks for traffic sign detection
  systems.
\newblock {\em Neurocomputing}, 316:332--344.

\bibitem[Athalye et~al., 2017]{athalye2017synthesizing}
Athalye, A., Engstrom, L., Ilyas, A., and Kwok, K. (2017).
\newblock Synthesizing robust adversarial examples.
\newblock {\em arXiv preprint arXiv:1707.07397}.

\bibitem[Carlini and Wagner, 2017]{carlini2017towards}
Carlini, N. and Wagner, D. (2017).
\newblock Towards evaluating the robustness of neural networks.
\newblock In {\em 2017 IEEE Symposium on Security and Privacy (SP)}, pages
  39--57. IEEE.

\bibitem[Chen et~al., 2018]{chen2018shapeshifter}
Chen, S.-T., Cornelius, C., Martin, J., and Chau, D. H.~P. (2018).
\newblock Shapeshifter: Robust physical adversarial attack on faster r-cnn
  object detector.
\newblock In {\em Joint European Conference on Machine Learning and Knowledge
  Discovery in Databases}, pages 52--68. Springer.

\bibitem[Eykholt et~al., 2018a]{eykholt2018detection}
Eykholt, K., Evtimov, I., Fernandes, E., Li, B., Rahmati, A., Tramer, F.,
  Prakash, A., Kohno, T., and Song, D. (2018a).
\newblock Physical adversarial examples for object detectors.
\newblock {\em arXiv preprint arXiv:1807.07769}.

\bibitem[Eykholt et~al., 2018b]{eykholt2018classification}
Eykholt, K., Evtimov, I., Fernandes, E., Li, B., Rahmati, A., Xiao, C.,
  Prakash, A., Kohno, T., and Song, D. (2018b).
\newblock Robust physical-world attacks on deep learning visual classification.
\newblock In {\em Proceedings of the IEEE Conference on Computer Vision and
  Pattern Recognition}, pages 1625--1634.

\bibitem[Goodfellow et~al., 2015]{FGSM}
Goodfellow, I., Shlens, J., and Szegedy, C. (2015).
\newblock Explaining and harnessing adversarial examples.
\newblock In {\em International Conference on Learning Representations}.

\bibitem[Jang et~al., 2017]{jang2017objective}
Jang, U., Wu, X., and Jha, S. (2017).
\newblock Objective metrics and gradient descent algorithms for adversarial
  examples in machine learning.
\newblock In {\em Proceedings of the 33rd Annual Computer Security Applications
  Conference}, pages 262--277.

\bibitem[Kingma and Welling, 2013]{kingma2013auto}
Kingma, D.~P. and Welling, M. (2013).
\newblock Auto-encoding variational bayes.
\newblock {\em arXiv preprint arXiv:1312.6114}.

\bibitem[Kurakin et~al., 2016]{kurakin2016adversarial}
Kurakin, A., Goodfellow, I., and Bengio, S. (2016).
\newblock Adversarial examples in the physical world.
\newblock {\em arXiv preprint arXiv:1607.02533}.

\bibitem[Liu et~al., 2016]{liu2016ssd}
Liu, W., Anguelov, D., Erhan, D., Szegedy, C., Reed, S., Fu, C.-Y., and Berg,
  A.~C. (2016).
\newblock Ssd: Single shot multibox detector.
\newblock In {\em European conference on computer vision}, pages 21--37.
  Springer.

\bibitem[Lu et~al., 2017]{lu2017standard}
Lu, J., Sibai, H., Fabry, E., and Forsyth, D. (2017).
\newblock Standard detectors aren't (currently) fooled by physical adversarial
  stop signs.
\newblock {\em arXiv preprint arXiv:1710.03337}.

\bibitem[Miyato et~al., 2018]{miyato2018spectral}
Miyato, T., Kataoka, T., Koyama, M., and Yoshida, Y. (2018).
\newblock Spectral normalization for generative adversarial networks.
\newblock {\em arXiv preprint arXiv:1802.05957}.

\bibitem[Moosavi-Dezfooli et~al., 2016]{moosavi2016deepfool}
Moosavi-Dezfooli, S.-M., Fawzi, A., and Frossard, P. (2016).
\newblock Deepfool: a simple and accurate method to fool deep neural networks.
\newblock In {\em Proceedings of the IEEE conference on computer vision and
  pattern recognition}, pages 2574--2582.

\bibitem[Ren et~al., 2015]{ren2015faster}
Ren, S., He, K., Girshick, R., and Sun, J. (2015).
\newblock Faster r-cnn: Towards real-time object detection with region proposal
  networks.
\newblock In {\em Advances in neural information processing systems}, pages
  91--99.

\bibitem[Rozsa et~al., 2016]{rozsa2016adversarial}
Rozsa, A., Rudd, E.~M., and Boult, T.~E. (2016).
\newblock Adversarial diversity and hard positive generation.
\newblock In {\em Proceedings of the IEEE Conference on Computer Vision and
  Pattern Recognition Workshops}, pages 25--32.

\bibitem[Sage et~al., 2018]{sage2018logo}
Sage, A., Agustsson, E., Timofte, R., and Van~Gool, L. (2018).
\newblock Logo synthesis and manipulation with clustered generative adversarial
  networks.
\newblock In {\em Proceedings of the IEEE Conference on Computer Vision and
  Pattern Recognition}, pages 5879--5888.

\bibitem[Sharif et~al., 2016]{sharif2016accessorize}
Sharif, M., Bhagavatula, S., Bauer, L., and Reiter, M.~K. (2016).
\newblock Accessorize to a crime: Real and stealthy attacks on state-of-the-art
  face recognition.
\newblock In {\em Proceedings of the 2016 acm sigsac conference on computer and
  communications security}, pages 1528--1540.

\bibitem[Sitawarin et~al., 2018]{sitawarin2018darts}
Sitawarin, C., Bhagoji, A.~N., Mosenia, A., Chiang, M., and Mittal, P. (2018).
\newblock Darts: Deceiving autonomous cars with toxic signs.
\newblock {\em arXiv preprint arXiv:1802.06430}.

\bibitem[Song et~al., 2018]{song2018unconstrained}
Song, Y., Shu, R., Kushman, N., and Ermon, S. (2018).
\newblock Constructing unrestricted adversarial examples with generative
  models.
\newblock In {\em Advances in Neural Information Processing Systems}, pages
  8312--8323.

\bibitem[Su et~al., 2019]{su2019one}
Su, J., Vargas, D.~V., and Sakurai, K. (2019).
\newblock One pixel attack for fooling deep neural networks.
\newblock {\em IEEE Transactions on Evolutionary Computation}, 23(5):828--841.

\bibitem[Szegedy et~al., 2013]{szegedy2013intriguing}
Szegedy, C., Zaremba, W., Sutskever, I., Bruna, J., Erhan, D., Goodfellow, I.,
  and Fergus, R. (2013).
\newblock Intriguing properties of neural networks.
\newblock {\em arXiv preprint arXiv:1312.6199}.

\bibitem[Tensorflow, 2020]{tensorflow_2020}
Tensorflow (2020).
\newblock tensorflow/docs.

\bibitem[Wang et~al., 2019]{wang2019nonconstrained}
Wang, X., He, K., Song, C., Wang, L., and Hopcroft, J.~E. (2019).
\newblock At-gan: An adversarial generator model for non-constrained
  adversarial examples.
\newblock {\em arXiv preprint arXiv:1904.07793}.

\bibitem[Xiao et~al., 2018]{xiao2018generating}
Xiao, C., Li, B., Zhu, J.-Y., He, W., Liu, M., and Song, D. (2018).
\newblock Generating adversarial examples with adversarial networks.
\newblock {\em arXiv preprint arXiv:1801.02610}.

\bibitem[Yang, 2018]{yang}
Yang, W. (2018).
\newblock Gan\_ lib\_tensorflow.

\bibitem[Yuan et~al., 2019]{yuan2019adversarial}
Yuan, X., He, P., Zhu, Q., and Li, X. (2019).
\newblock Adversarial examples: Attacks and defenses for deep learning.
\newblock {\em IEEE transactions on neural networks and learning systems},
  30(9):2805--2824.

\bibitem[Zhao et~al., 2017]{zhao2017generating}
Zhao, Z., Dua, D., and Singh, S. (2017).
\newblock Generating natural adversarial examples.
\newblock {\em arXiv preprint arXiv:1710.11342}.

\end{thebibliography}

\end{document}